\documentclass[twocolumn]{autart}    

\usepackage{mathtools}
\usepackage{amssymb}
\usepackage{amsmath}
\usepackage{siunitx}
\usepackage{tikz}
\usepackage{pgfplots}
\usepackage{float}
\usepackage[numbers]{natbib}
\usepgfplotslibrary{fillbetween}
\usepackage{graphicx}
\usepackage{setspace}
\usepackage{booktabs}
\usepackage{url}
\usepackage{multicol}

\let\classAND\AND
\let\AND\relax
\usepackage{algorithm}
\usepackage{algorithmic}

\let\AND\classAND
\AtBeginEnvironment{algorithmic}{\let\AND\algoAND}

\floatname{algorithm}{Algorithm}

\DeclareMathAlphabet\mathbfcal{OMS}{cmsy}{b}{n}
\newcommand{\K}{\mathbf{K}}

\newcommand{\Pp}{\mathbf{P}}

\newcommand{\Ll}{\mathbf{L}}

\newcommand{\Ss}{\mathbf{S}}

\newcommand{\x}{\mathbf{x}}

\newcommand{\w}{\mathbf{w}}

\newcommand{\y}{\mathbf{y}}

\newcommand{\id}{\mathbf{I}}

\definecolor{blue}{RGB}{3, 128, 149}
\definecolor{green}{RGB}{59, 151, 52}
\definecolor{red}{RGB}{248, 60, 93}
\definecolor{yellow}{RGB}{247, 179, 51}
\definecolor{lightred}{RGB}{253, 223, 223}
\definecolor{lightblue}{RGB}{222, 243, 253}
\definecolor{lightgreen}{RGB}{191, 216, 213}
\definecolor{lightyellow}{RGB}{246, 234, 197}

\begin{document}

\begin{frontmatter}
\title{Tensor network square root Kalman filter \\ for online Gaussian process regression} 
\author[Delft]{Clara Menzen}\ead{cm.menzen@gmail.com},
\author[Delft]{Manon Kok}\ead{m.kok-1@tudelft.nl},
\author[Delft]{Kim Batselier}\ead{k.batselier@tudelft.nl}

\address[Delft]{Delft Center for Systems and Control, Delft University of Technology, Mekelweg 2, 2628 CD, Delft, the Netherlands}

                                        
\begin{abstract}
The state-of-the-art tensor network Kalman filter lifts the curse of dimensionality for high-dimensional recursive estimation problems.
However, the required rounding operation can cause filter divergence due to the loss of positive definiteness of covariance matrices.
We solve this issue by developing, for the first time, a tensor network square root Kalman filter, and apply it to high-dimensional online Gaussian process regression.
In our experiments, we demonstrate that our method is equivalent to the conventional Kalman filter when choosing a full-rank tensor network.
Furthermore, we apply our method to a real-life system identification problem where we estimate $4^{14}$ parameters on a standard laptop.
The estimated model outperforms the state-of-the-art tensor network Kalman filter in terms of prediction accuracy and uncertainty quantification.
\end{abstract}

\end{frontmatter}
\section{Introduction}
In a time when data-driven AI models are trained on an exponentially growing amount of data, it is crucial that the models can be adapted to newly observed data without retraining from scratch.
These online or recursive settings are present in many fields including system identification \cite{batselier2017TNKF,doyle2002identification}, sensor fusion \cite{viset2022extended,solin2018modeling}, robotics \cite{nguyen2008local,Miao}, and machine learning \cite{hartikainen2010kalman,ranganathan2010online,stanton2021kernel}.

While Bayesian algorithms, like widely-used Gaussian processes (GPs)  \cite{Rasmussen2006} are well-suited for an online setting, they are associated with potentially high computational costs.
Standard GP regression using a batch of $N$ observations has a cubic cost in $N$, i.e., $\mathcal{O}(N^3)$.
The number of observations is growing in an online setting, so the cost increases each time step and can become a computational bottleneck.

There are numerous parametric approximations to address scalability in batch settings, including sparse GPs \cite{quinonero2005unifying} and reduced-rank GPs \cite{HilbertGP}, which both have a complexity of $\mathcal{O}(NM^2)$, $M$ being the number of inducing inputs and basis function for the respective method.
Structured kernel interpolation for sparse GPs \cite{wilson2015kernel} reduces the complexity further to $\mathcal{O}(N+DM^{1+1/D})$, $D$ being the number of input dimensions.

Parametric approximations allow for a straightforward recursive update, where the posterior distribution from the previous time step is used as a prior for the current time step \cite{sarkka2023bayesian}.
In this context, online GPs have been used, e.g., \ for GP state-space models \cite{sarkka2013spatiotemporal,svensson2017flexible,berntorp2021online}, rank-reduced Kalman filtering \cite{schmidt2023rank} and recursive sparse GPs \cite{stanton2021kernel}.

In this paper, we consider the online parametric GP model given by
\begin{equation}
\begin{aligned}
    y_t =  \boldsymbol{\phi}&(\x_t)^\top \w_t + \epsilon_t,\quad\epsilon_t\sim\mathcal{N}(0,\sigma_y^2),\\
    &\w_{t-1}\sim\mathcal{N}(\hat\w_{t-1},\mathbf{P}_{t-1}),
\end{aligned}
    \label{eq:parammodel}
\end{equation}
where $y_t$ is a scalar observation at discrete time $t$, $\boldsymbol{\phi}(\cdot)$ are basis functions that map a $D$ dimensional input vector $\x_t$ to a feature space, $\w_t\in\mathbb{R}^M$ are the parameters at time $t$, and $\sigma_y^2$ denotes the variance of the measurement noise $\epsilon_t$ which is assumed to be i.d.d. and zero-mean Gaussian.
With \eqref{eq:parammodel}, the posterior distribution \mbox{$p(\w_t\mid\y_{1:t})$} is computed each time step using the estimate $\hat\w_{t-1}$ and covariance matrix $\Pp_{t-1}$ of the previous time step as a prior. 

We consider commonly used product kernels with a feature map given by
\begin{equation}
    \boldsymbol{\phi}(\x_t) = \boldsymbol{\phi}^{(1)}(\x_t) \otimes \cdots \otimes\boldsymbol{\phi}^{(d)}(\x_t) \otimes\cdots\otimes\boldsymbol{\phi}^{(D)}(\x_t), \label{eq:featuremap}
\end{equation}
where $\boldsymbol{\phi}^{(d)}(\x_t)\in\mathbb{R}^I$ with $I$ being the number of basis functions in the $d$th dimension.
The resulting number of basis functions is $M=I^D$, growing exponentially with the input dimension $D$.
Several tensor network (TN)- based methods have been proposed to break this curse of dimensionality and achieve a linear computational complexity in $D$.
In the batch setting, \cite{batselier2017VoleterrALS} and \cite{wesel2021large} give solutions for the squared exponential and polynomial kernel, respectively.
In the online setting, the state-of-the-art method is the tensor network Kalman filter (TNKF) \cite{batselier2017TNKF, Batselier2019extended}, where the Kalman filter time and measurement update are implemented in TN format. 

While the TNKF lifts the curse of dimensionality, it has a significant drawback.
The TNKF requires a TN-specific rounding operation \cite{Oseledets2011}, which can result in covariance update losing positive  \mbox{(semi-)} definiteness \cite{de2023enabling}, resulting in the divergence of the filter.


This paper resolves this issue by computing the square root covariance factor in tensor train (TT) format instead.
Our approximation represents the $M\times M$ square root covariance factor as a tensor train matrix (TTm). 
This is motivated by prior square root covariance factors of product kernels having a Kronecker product structure, which corresponds to a rank-1 TTm.
In addition, work by \cite{nickson2015blitzkriging} and \cite{izmailov2018scalable} approximate the covariance matrix as a rank-1 TTm.
This work generalizes the rank-1 approximation to higher ranks which results in better prediction accuracy and uncertainty quantification.
We call our method the tensor network square root Kalman filter (TNSRKF).

We show in experiments that the TNSRKF is equivalent to the standard Kalman filter when choosing full-rank TTs.
In addition, we show how different choices of TT-ranks affect the performance of our method.
Finally, we compare the TNSRKF to the TNKF in a real life system identification problem with $4^{14}$ parameters, and show that contrary to the TNKF, our method does not diverge.

\section{Problem Formulation}
Similar to the TNKF, we build on standard equations for the measurement update of the Kalman filter given by
\begin{align}
    \Ss_t &= \boldsymbol\phi_t^\top \Pp_{t-1} \boldsymbol\phi_t + \sigma_y^2\label{eq:innovation}\\
    \K_t &= \Pp_{t-1}\boldsymbol\phi_t \Ss_t^{-1}\label{eq:Kalmangain}\\
    \hat\w_t &= \hat\w_{t-1} + \K_t(y_t - \boldsymbol\phi_t^\top \hat\w_{t-1})\label{eq:meanupdate}\\
    \Pp_t &= (\id_M-\K_t\boldsymbol\phi_t^\top)\Pp_{t-1}(\id_M-\K_t\boldsymbol\phi_t^\top)^\top + \sigma_y^2\K_t\K^\top_t,\label{eq:covaupdate}
\end{align}%
where $\Ss_t$ denotes the innovation covariance and $\K_t$ denotes the Kalman gain. 
Note that for a scalar measurement, $\Ss_t$ is a scalar and $\K_t$ a vector, whereas in the case of multiple measurements per time step, they are matrices.
In this way, we recursively update the posterior distribution of the parametric weights from \eqref{eq:parammodel}, i.e., \ $p(\w_t\mid\y_{1:t})$.
For product kernels with a feature map given in \eqref{eq:featuremap}, it is $\w_t\in\mathbb{R}^{I^D}$ and $\Pp_t\in\mathbb{R}^{I^D\times I^D}$.
In this case, the Kalman filter suffers from the curse of dimensionality.

The first tensor-based Kalman filter, the TNKF \cite{batselier2017TNKF}, solved the curse of dimensionality and implements \eqref{eq:innovation}-\eqref{eq:covaupdate} in TT format, where the weights are represented as a TT and the covariance matrix as a TTm.
During the updates, the algebraic operations in TT format increase the TT-ranks of the involved variables, according to \cite[Lemma 2]{Batselier2019extended}.
To counteract the rank increase and keep the algorithm efficient, the TNKF requires an additional step called TT-rounding \cite{Oseledets2011}.
This SVD-based operation transforms the TT or TTm to ones with smaller TT-ranks.
TT-rounding can result, however, in the loss of positive (semi-) definiteness.

To avoid this issue, we implement the square root formulation of the Kalman filter (SRKF), as described e.g.\ in \cite[Ch. 7]{grewal2014kalman}, in TT format.
The SRKF expresses \eqref{eq:innovation}-\eqref{eq:covaupdate} in terms of the square root covariance factor, given by
\begin{equation}
    \Ll_t = \begin{bmatrix} (\id_M-\K_t\boldsymbol\phi_t^\top)\Ll_{t-1}  & \;\;\; \sigma_y\K_t \end{bmatrix},\label{eq:sqrt}
\end{equation}
which consists of a concatenating matrices that increases the number of columns.
For the next update, $\Ll_t$ needs to be transformed back to its original size.
In the SRKF, this is done by computing a thin QR-decomposition \cite[p. 248]{golub2013matrix} of $\Ll_t$ given by
\begin{equation}
    \underbrace{\Ll_t^\top}_{(M+1)\times M} = \underbrace{\mathbf{Q}_t}_{(M+1)\times M}\underbrace{\mathbf{R}_t}_{M\times M}\label{eq:augL}
\end{equation}
and replacing $\Ll_t$ by $\mathbf{R}_t$.
The orthogonal $\mathbf{Q}_t$-factor can be discarded since
\begin{equation}
    \Pp_t = \Ll_t\Ll_t^\top = \mathbf{R}_t^\top\underbrace{\mathbf{Q}_t^\top\mathbf{Q}_t}_{\mathbf{I}_M}\mathbf{R}_t = \mathbf{R}_t^\top\mathbf{R}_t.
    \label{eq:covfromQR}
\end{equation}

In TT format, performing the QR-decomposition as in \eqref{eq:augL} is not possible.
We solve this issue by proposing an SVD-based algorithm in TT format that truncates $\Ll_t$ back to its original size.


\section{Background on tensor networks}
\subsection{Tensor networks}
Tensor networks (TNs), also called tensor decompositions, are an extension of matrix decompositions to higher dimensions.
In this paper, we use a specific architecture of TNs, called TTs \cite{Oseledets2011} to approximate the weight vector's mean as discussed in Section \ref{sec:TTv}, and a TT matrix (TTm) \cite{Oseledets2010} to approximate the square root covariance factor, as discussed in Section \ref{sec:TTm}.

In this context, we denote TTs representing vectors as a lower-case bold letter, e.g.\ $\w_t$, and their components, called TT-cores, as capital calligraphic bold letters, e.g.\ $\mathbfcal{W}^{(d)}$.
TT matrices are denoted by upper-case bold letters, e.g.\ $\Ll_t$ and their corresponding TTm-cores as capital calligraphic bold letters, e.g.\ $\mathbfcal{L}^{(d)}$.

\begin{figure}
    \centering
    \begin{tikzpicture}
\draw  (-0.9,1.8) ellipse (0.2 and 0.2);
\draw  (-0.9,1.2) ellipse (0.2 and 0.2);
\draw  (-0.9,0.6) ellipse (0.2 and 0.2);
\draw  (-0.9,0) ellipse (0.2 and 0.2);
\draw  (-0.9,-0.6) ellipse (0.2 and 0.2);
\draw (-0.9,1.6) -- (-0.9,1.4);
\draw (-0.9,1) -- (-0.9,0.8);
\draw (-0.9,0.4) -- (-0.9,0.2);
\draw (-0.9,-0.2) -- (-0.9,-0.4);
\draw [red] (-1.1,1.8) -- (-1.3,1.8);
\draw [red] (-1.1,1.2) -- (-1.3,1.2);
\draw [red] (-1.1,0.6) -- (-1.3,0.6);
\draw [red] (-1.1,0) -- (-1.3,0);
\draw [red] (-1.1,-0.6) -- (-1.3,-0.6);

\draw  (-2.4,1.8) ellipse (0.2 and 0.2);
\draw  (-2.4,1.2) ellipse (0.2 and 0.2);
\draw  (-2.4,0.6) ellipse (0.2 and 0.2);
\draw  (-2.4,0) ellipse (0.2 and 0.2);
\draw  (-2.4,-0.6) ellipse (0.2 and 0.2);
\draw  (-2.4,1.6) -- (-2.4,1.4);
\draw (-2.4,1) -- (-2.4,0.8);
\draw (-2.4,0.4) -- (-2.4,0.2);
\draw (-2.4,-0.2) -- (-2.4,-0.4);
\draw (-2.6,1.8) -- (-2.8,1.8);
\draw (-2.6,1.2) -- (-2.8,1.2);
\draw (-2.6,0.6) -- (-2.8,0.6);
\draw (-2.6,0) -- (-2.8,0);
\draw (-2.6,-0.6) -- (-2.8,-0.6);
\draw [blue] (-0.7,1.8) -- (-0.5,1.8);
\draw [blue] (-0.7,1.2) -- (-0.5,1.2);
\draw [blue] (-0.7,0.6) -- (-0.5,0.6);
\draw [blue] (-0.7,0) -- (-0.5,0);
\draw [blue] (-0.7,-0.6) -- (-0.5,-0.6);

\draw  (0.7,1.8) ellipse (0.2 and 0.2);
\draw  (0.7,1.2) ellipse (0.2 and 0.2);
\draw  (0.7,0.6) ellipse (0.2 and 0.2);
\draw  (0.7,0) ellipse (0.2 and 0.2);
\draw  (0.7,-0.6) ellipse (0.2 and 0.2);
\draw (0.7,1.6) -- (0.7,1.4);
\draw (0.7,1) -- (0.7,0.8);
\draw (0.7,0.4) -- (0.7,0.2);
\draw (0.7,-0.2) -- (0.7,-0.4);
\draw (0.5,1.8) -- (0.3,1.8);
\draw (0.5,1.2) -- (0.3,1.2);
\draw (0.5,0.6) -- (0.3,0.6);
\draw (0.5,0) -- (0.3,0);
\draw (0.5,-0.6) -- (0.3,-0.6);
\draw (0.9,0.6) -- (1.2,0.6);

\draw  (2,1.8) ellipse (0.2 and 0.2);
\draw  (2,1.2) ellipse (0.2 and 0.2);
\draw  (2,0.6) ellipse (0.2 and 0.2);
\draw  (2,0) ellipse (0.2 and 0.2);
\draw  (2,-0.6) ellipse (0.2 and 0.2);
\draw (2,1.6) -- (2,1.4);
\draw (2,1) -- (2,0.8);
\draw (2,0.4) -- (2,0.2);
\draw (2,-0.2) -- (2,-0.4);
\draw (1.8,1.8) -- (1.6,1.8);
\draw (1.8,1.2) -- (1.6,1.2);
\draw (1.8,0.6) -- (1.6,0.6);
\draw (1.8,0) -- (1.6,0);
\draw (1.8,-0.6) -- (1.5,-0.6);
\draw (2.2,0.6) -- (2.4,0.6);
\draw  (2.6,0.6) ellipse (0.2 and 0.2);
\draw (2.8,0.6) -- (2.94,0.6);
\draw  (3.2,0.6) ellipse (0.25 and 0.2);
\draw (3.43,0.6) -- (3.75,0.6);
\node at (-2.4,-1.03) {\tiny TT};
\node at (-0.9,-1.03) {\tiny TTm};
\node at (0.7,-1.03) {\tiny tall TTm};
\node at (2.1,-0.93) {\tiny thin SVD in};
\node at (2.1,-1.13) {\tiny TT format};
\node [red,rotate=-90] at (-1.5,0.6) {\tiny row indices};
\node [blue,rotate=-90] at (-0.3,0.6) {\tiny column indices};
\node at (2.6,0.6) {\tiny$\mathbf{S}$};
\node at (3.2,0.6) {\tiny$\mathbf{V}^\top$};
\node at (-2.4,-1.33) {\tiny (a)};
\node at (-0.9,-1.33) {\tiny (b)};
\node at (0.7,-1.33) {\tiny (c)};
\node at (2,-1.33) {\tiny (d)};
\end{tikzpicture}
    \caption{Visual depiction of tensor diagrams for a (a) TT, (b) TTm, (c) tall TTm and (d) thin SVD.}
    \label{fig:thinTT}
\end{figure}
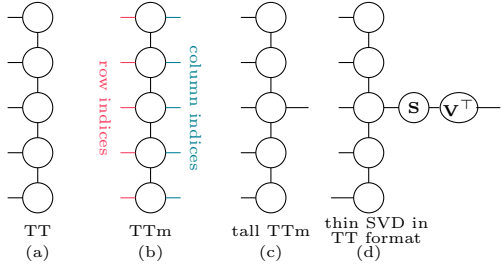

\subsubsection{Tensor train vectors}
\label{sec:TTv}
As depicted in Fig.\ \ref{fig:thinTT}(a), a TT vector consists of interconnected three-way tensors, called TT-cores, visualized as nodes with three edges.
Each edge corresponds to an index of a TT-core and connected edges are summations over the involved indices.
Each TT-core is connected by two edges, called TT-ranks, to their neighbouring TT-cores, except for the first and last TT-core, which outer TT-ranks are by definition equal to one.

For the purpose of this paper, consider a TT that represents the mean of the weight vector $\w_t\in\mathbb{R}^M$.
The TT-cores, denoted by $\mathbfcal{W}_t^{(1)},\cdots,\mathbfcal{W}_t^{(d)},\cdots,\mathbfcal{W}_t^{(D)}$ are each of size $R_d\times I\times R_{d+1}$, where $R_d$ and $R_{d+1}$ are the TT-ranks and $I$ is the size of the non-connected edge such that \mbox{$M=I^D$}.
By definition $R_1=R_{D+1}=1$.
Without the loss of generality, we use TT-cores with equal TT-ranks $R_\w$.
The storage complexity of $\w_t$ without TNs is $\mathcal{O}(I^D)$ and in TT format $\mathcal{O}(DIR_\w^2)$, where lower TT-ranks $R_\w$ will result in more efficient representations.

An important characteristic of a TT for numerical stability is that it can be transformed into the site-$d$-mixed canonical format.
\begin{defn}\textbf{Site-$d$-mixed canonical format \cite{Schollwock2011}}
    A TT $\w_t$ in site-$d$-mixed canonical format is given by 
    \begin{equation}
    \w_t=\mathbf{G}_{d,t}\w_t^{(d)},    \label{eq:sited_w}
    \end{equation}
    where $\mathbf{G}_{d,t}\in\mathbb{R}^{M\times R_\w IR_\w}$ is an orthogonal matrix computed from all TT-cores except the $d$th and \mbox{$\w_t^{(d)}\in\mathbb{R}^{R_\w IR_\w}$} is the vectorization of the $d$th TT-core.
    In this format, the TT representation is linear in the $d$th TT-core when all other TT-cores are fixed. \label{def:sitek}
\end{defn}

\subsubsection{TT matrices and tall TT matrices}
\label{sec:TTm}
A TTm consists of interconnected four-way tensors, as depicted in Fig.\ \ref{fig:thinTT}(b).
Analogous to the TT, the TTm components and connected edges are called TTm-cores and TTm-ranks, respectively, where each TTm-core has two free edges, the row and column indices.

For the purpose of this paper, consider a TTm representation of the square root covariance factor $\Ll_t\in\mathbb{R}^{M\times M}$.
The TTm-cores $\mathbfcal{L}_t^{(1)},\cdots,\mathbfcal{L}_t^{(d)},\cdots,\mathbfcal{L}_t^{(D)}$ are of size $R_d\times I\times J \times R_{d+1}$, where $I$ and $J$ are the row and column indices, indicated in Fig.\ \ref{fig:thinTT}(b) as red and and blue edges respectively, such that \mbox{$M=I^D$} and $M=J^D$.
By definition, $R_1=R_{D+1}=1$, and for this paper, we generally assume that all other TTm-ranks $R_2=\cdots=R_D=R_\Ll$ are equal.
The storage complexity of $\Ll_t$ without TNs is $\mathcal{O}(I^D\times I^D)$ and in TTm format $\mathcal{O}(DR_\Ll^2 IJ)$.

A TTm can also be written in terms of the site-$d$-mixed canonical format as defined in Definition \ref{def:sitek}, but it requires to be transformed into a TT first.
This can be done by combining the row and column indexes into one index, which represents a kind of vectorization of the matrix represented by the TTm. 
Note, however, that the indices are not ordered as in conventional vectorization.
A site-$d$-mixed canonical format of a TTm is given by 
\begin{equation}
 \operatorname{vec}(\Ll_t) =\mathbf{H}_{d,t}\mathbf{l}_t^{(d)},
 \label{eq:sitek_ttm}
\end{equation}
where $\mathbf{H}_{d,t}\in\mathbb{R}^{2M\times R_\Ll IJR_\Ll}$ is computed from all the TTm-cores but the $d$th, and $\mathbf{l}^{(d)}_t\in\mathbb{R}^{R_\Ll IJR_\Ll}$.

To recompute $\Ll_t$ back in its original like in the QR step of the SRKF (see \eqref{eq:augL}), we need a special of a TTm, the tall TTm, as well as a thin SVD in TTm format.
\begin{defn}\textbf{Tall TTm \cite{batselier2017VoleterrALS}}
    A tall TTm, as depicted in Figure \ref{fig:thinTT}(c), has only one TTm-core with both a row and column index, while all other TTm-cores have only row indices.
    Then, the TTm represents a tall matrix with many more rows than columns.
\end{defn}

\begin{defn}\textbf{Thin SVD in TTm format \cite{batselier2022low}}
\label{def:svdthin}
   Consider a TTm in site-$d$-mixed canonical format, where the $d$th TTm-core is the one that has the column index, \mbox{$\mathbfcal{L}^{(d)}\in\mathbb{R}^{R_\Ll\times I\times J\times R_\Ll}$}.
   The SVD of $\mathbfcal{L}^{(d)}$ reshaped and permuted in to a matrix of size $R_\Ll IR_\Ll\times J$, is given by
   \begin{equation}
       \mathbf{U}^{(d)}\mathbf{S}^{(d)}(\mathbf{V}^{(d)})^\top.
   \end{equation}
    Now replace the $d$th TTm-core by $\mathbf{U}^{(d)}$ reshaped and permuted back to the original TTm-core dimensions.  
    
   Then the thin SVD is given by the TTm with the replaced TT-core as the orthogonal $\mathbf{U}$-factor, and $\mathbf{S}^{(d)}(\mathbf{V}^{(d)})^\top$ as the $\mathbf{SV}^\top$-factors, as depicted in Fig.\ \ref{fig:thinTT}(d).
\end{defn}

\section{Tensor-networked SRKF}
We propose our method combining efficient TN methods with the SRKF formulation for online GP regression.
More specifically, we recursively compute the posterior distribution of the parametric weights in \eqref{eq:parammodel} from the measurement update of the Kalman filter.
To achieve this, we update the mean $\hat\w_t\in\mathbb{R}^M$ as a TT (Section \ref{sec:meanupdate}), and the square root factor $\Ll_t\in\mathbb{R}^{M\times M}$ as a TTm (Section \ref{sec:srupdate}).

All computations are summarized in Algorithm \ref{alg:onlineregression}, which outputs the posterior weight distributions $p(\w_t\mid\y_{1:t})=\mathcal{N}(\hat{\w}_t,\Pp_t)$, and the predictive distributions for a test input $p(f_{*,t})=\mathcal{N}(m_{*,t},\sigma_{*,t}^2)$.

\subsection{Update of weight mean}
\label{sec:meanupdate}
For updating the mean with a new measurement $y_t\in\mathbb{R}$, we compute \eqref{eq:meanupdate} in TT format.
In the original tensor-based KF \cite{batselier2017TNKF}, the two terms in equation \eqref{eq:meanupdate} are summed together in TT format, which increases the TT-ranks.
To avoid this rank increase, we apply a commonly-used optimization algorithm from the tensor community, called the alternating linear scheme (ALS) \cite{Rohwedder2012,Rohwedder2013}.
The ALS computes a TT by updating one TT-core at a time while keeping all other TT-cores fixed.
The optimization problem to be solved is given by
\begin{equation}
\begin{aligned}
    \underset{\w_t}{\operatorname{min}} &\| \hat\w_{t-1} + \K_t(y_t-\boldsymbol\phi_t^\top \hat\w_{t-1}) - \w_t \|_\mathrm{F}^2\\
    & \text{s.t.}\;\; \w_t\;\;\text{being a low-rank TT}, 
    \label{eq:optproblem}
\end{aligned}
\end{equation}
where the subscript F stands for the Frobenius norm, and $\hat\w_{t-1}$ is the estimate from the last time step playing now the role of the prior for the current time step.

Inserting \eqref{eq:sited_w} in \eqref{eq:optproblem}, thus making use of the site-$d$-mixed canonical format from Definition \ref{def:sitek}, gives the optimization problem for the update of one TT-core 
\begin{equation}
    \underset{\w_t^{(d)}}{\operatorname{min}} \left\| \mathbf{G}_{d,t}^\top\left(\hat\w_{t-1} + \mathbf{K}_t(y_t-\boldsymbol\phi_t^\top \hat\w_{t-1})\right) - \w_t^{(d)} \right\|_\mathrm{F}^2. \label{eq:ALS}
\end{equation}
In one so-called sweep of the ALS, \eqref{eq:ALS} is solved for each TT-core once.
A stopping criterion for the convergence of the residual in \eqref{eq:ALS} determines the total number of sweeps.

\subsection{Update of square root covariance factor}
\label{sec:srupdate}
To compute the covariance matrix with the standard covariance update in the measurement update, see \eqref{eq:covaupdate}, we recursively compute the square root covariance factor $\Ll_t$ as defined in \eqref{eq:sqrt} such that $\Pp_t=\Ll_t\Ll_t^\top$.
To achieve this, we use the ALS to solve \eqref{eq:sqrt} (ALS step) and then we transform $\Ll_t$ as in \eqref{eq:augL} back to its original size (QR step).

\paragraph*{ALS step}
In this step, we use the ALS to compute a TTm representing $\Ll_t$. 
We solve the optimization problem given by
\begin{equation}
    \begin{aligned}
        \min_{\Ll_t}& \left\| 
        \begin{bmatrix} (\id_M-\K_t\boldsymbol\phi_t^\top)\Ll_{t-1} &\;\; \sigma_y\K_t \end{bmatrix} - \Ll_t \right\|_\mathrm{F}^2 \\
    & \text{s.t.}\;\; \Ll_t\;\;\text{being a low-rank TTm,}
    \label{eq:covaopt}
    \end{aligned}
\end{equation}
where $\Ll_{t-1}$ is the estimated square root covariance factor from time step $t-1$ now serving as the prior.
The original ALS algorithm is defined for TTs, so we must adapt it for TT matrices.

For this, it is necessary to use the site-$d$-mixed canonical form for TT matrices, as described in Section \ref{sec:TTm} above \eqref{eq:sitek_ttm}.
In addition, we need to horizontally concatenate two matrices in TTm format, which can be done by summing two matrices of size $M\times 2M$ such that \eqref{eq:covaopt} becomes
\begin{equation}
\begin{aligned}
    \min_{\mathbf{l}_t^{(d)}} \left\| \mathbf{H}_{d,t}^\top \operatorname{vec}\left(\begin{bmatrix}1&0\end{bmatrix} \otimes (\id_M-\K_t\boldsymbol\phi_t^\top)\Ll_{t-1}\right) \right. \\
     \left.+\mathbf{H}_{d,t}^\top\operatorname{vec}\left(\begin{bmatrix}0& 1\end{bmatrix}\otimes \begin{bmatrix}1&\boldsymbol0_{M-1}\end{bmatrix}\otimes\sigma_y\K_t\right)   - \mathbf{l}_t^{(d)}) \right\|_\mathrm{F}^2,\label{eq:ALScov}
\end{aligned}
\end{equation}
where vec denotes the vectorizations of the involved TT matrices.

\paragraph*{QR step}
The optimization problem given by \eqref{eq:covaopt} requires concatenating a matrix with a column vector.
In TT format, this results in a TTm of size $M\times 2M$.
For the TTm-cores of $\Ll_t$ this means that one TTm-core, we call it the augmented core, is of size $R_\Ll \times I\times 2J\times R_\Ll$.
Before serving as a prior for the next time step, a QR step as in \eqref{eq:augL} is required to transform $\Ll_t$ back to its original size.
We use an SVD-based algorithm in TN format to transform $\Ll_t$ of size $M\times 2M$ back to $M\times M$, as described in Algorithm \ref{alg:retruncate}.

\subsection{Predictions}
To perform GP predictions we compute the predictive distribution for a test output $f_{*,t}=\boldsymbol{\phi}(\x_{*})^\top\w_t$ with mean and variance given by
\begin{equation}
\begin{aligned}
    m_{*,t} &= \boldsymbol\phi(\x_{*})^\top\hat\w_t\\
    \sigma^2_{*,t} &= \boldsymbol\phi(\x_{*})^\top\Ll_t\Ll_t^\top\boldsymbol\phi(\x_{*}).
    \label{eq:predpost}
\end{aligned}
\end{equation}

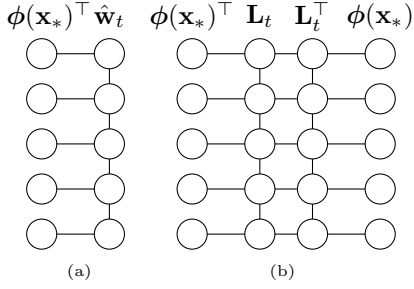
\begin{figure}
    \centering
    \begin{tikzpicture}
\draw  (1.4,1.1) ellipse (0.2 and 0.2);
\draw  (1.4,0.5) ellipse (0.2 and 0.2);
\draw  (1.4,-0.1) ellipse (0.2 and 0.2);
\draw  (1.4,-0.7) ellipse (0.2 and 0.2);
\draw  (1.4,-1.3) ellipse (0.2 and 0.2);
\draw  (2.3,1.1) ellipse (0.2 and 0.2);
\draw  (2.3,0.5) ellipse (0.2 and 0.2);
\draw  (2.3,-0.1) ellipse (0.2 and 0.2);
\draw  (2.3,-0.7) ellipse (0.2 and 0.2);
\draw  (2.3,-1.3) ellipse (0.2 and 0.2);
\draw (2.3,0.9) -- (2.3,0.7);
\draw (2.3,0.3) -- (2.3,0.1);
\draw (2.3,-0.3) -- (2.3,-0.5);
\draw (2.3,-0.9) -- (2.3,-1.1);
\draw (2.1,1.1) -- (1.6,1.1);
\draw (2.1,0.5) -- (1.6,0.5);
\draw (2.1,-0.1) -- (1.6,-0.1);
\draw (2.1,-0.7) -- (1.6,-0.7);
\draw (2.1,-1.3) -- (1.6,-1.3);
\node at (1.5,1.6) {$\boldsymbol{\phi}(\mathbf{x}_*)^\top$};
\node at (2.3,1.6) {$\hat{\mathbf{w}}_t$};
\draw  (3.4,1.1) ellipse (0.2 and 0.2);
\draw  (3.4,0.5) ellipse (0.2 and 0.2);
\draw  (3.4,-0.1) ellipse (0.2 and 0.2);
\draw  (3.4,-0.7) ellipse (0.2 and 0.2);
\draw  (3.4,-1.3) ellipse (0.2 and 0.2);
\draw  (4.3,1.1) ellipse (0.2 and 0.2);
\draw  (4.3,0.5) ellipse (0.2 and 0.2);
\draw  (4.3,-0.1) ellipse (0.2 and 0.2);
\draw  (4.3,-0.7) ellipse (0.2 and 0.2);
\draw  (4.3,-1.3) ellipse (0.2 and 0.2);
\draw (4.3,0.9) -- (4.3,0.7);
\draw (4.3,0.3) -- (4.3,0.1);
\draw (4.3,-0.3) -- (4.3,-0.5);
\draw (4.3,-0.9) -- (4.3,-1.1);
\draw (4.1,1.1) -- (3.6,1.1);
\draw (4.1,0.5) -- (3.6,0.5);
\draw (4.1,-0.1) -- (3.6,-0.1);
\draw (4.1,-0.7) -- (3.6,-0.7);
\draw (4.1,-1.3) -- (3.6,-1.3);
\node at (3.4,1.6) {$\boldsymbol{\phi}(\mathbf{x}_*)^\top$};
\node at (4.3,1.6) {$\mathbf{L}_t$};
\draw  (5,1.1) ellipse (0.2 and 0.2);
\draw  (5,0.5) ellipse (0.2 and 0.2);
\draw  (5,-0.1) ellipse (0.2 and 0.2);
\draw  (5,-0.7) ellipse (0.2 and 0.2);
\draw  (5,-1.3) ellipse (0.2 and 0.2);
\draw  (5.9,1.1) ellipse (0.2 and 0.2);
\draw  (5.9,0.5) ellipse (0.2 and 0.2);
\draw  (5.9,-0.1) ellipse (0.2 and 0.2);
\draw  (5.9,-0.7) ellipse (0.2 and 0.2);
\draw  (5.9,-1.3) ellipse (0.2 and 0.2);
\node at (5,1.6) {$\mathbf{L}_t^\top$};
\node at (5.9,1.6) {$\boldsymbol{\phi}(\mathbf{x}_*)$};
\draw (4.5,1.1) -- (4.8,1.1);
\draw (4.5,0.5) -- (4.8,0.5);
\draw (4.5,-0.1) -- (4.8,-0.1);
\draw (4.5,-0.7) -- (4.8,-0.7);
\draw (4.5,-1.3) -- (4.8,-1.3);
\draw (5.2,1.1) -- (5.7,1.1);
\draw (5.2,0.5) -- (5.7,0.5);
\draw (5.2,-0.1) -- (5.7,-0.1);
\draw (5.2,-0.7) -- (5.7,-0.7);
\draw (5.2,-1.3) -- (5.7,-1.3);
\draw (5,0.9) -- (5,0.7);
\draw (5,0.3) -- (5,0.1);
\draw (5,-0.3) -- (5,-0.5);
\draw (5,-0.9) -- (5,-1.1);
\node at (1.9,-1.8) {\tiny(a)};
\node at (4.6,-1.8) {\tiny(b)};
\end{tikzpicture}
    \caption{Visual depiction of (a) predictive mean and (b) predictive covariance for $D=5$.}
    \label{fig:pred}
\end{figure}

Given $\hat\w_t$ as a TT and $\Ll_t$ as a TTm, we can compute \eqref{eq:predpost} directly in TN format without explicitly reconstructing the mean vector and square root factor.
For a test input $\x_*$, Fig.\ \ref{fig:pred} illustrates the computation of (a) the predictive mean $m_{*,t}$, (b) the predictive covariance $\sigma^2_{*,t}$.

\begin{algorithm}
\caption{Online GP regression in terms of SRKF in TT format (TNSRKF)} \label{alg:onlineregression}
    \begin{algorithmic}[1]\onehalfspacing
        \REQUIRE Measurements $\y=y_1,y_2,\dots,y_N$, \\
        basis functions for inputs $\boldsymbol{\phi}(\x_t)$, $t=1,\dots,N$, \\ 
        prior $\hat\w_0$ in TN format \mbox{(Lemma \ref{lem:zmprior})}\\
        prior $\Ll_0$ in TN format (Lemma \ref{lem:srcovM}), \\ 
        noise variance $\sigma_y^2$, \\
        basis functions for prediction point $\boldsymbol{\phi}(\x_{*})$.
        \ENSURE $p(\w\mid\y_{1:t})=\mathcal{N}(\hat\w_t,\Pp_t)$ and \\ $p(f_{*}\mid\y_{1:t})=\mathcal{N}(m_{*,t},\sigma^2_{*,t})$, for $t=1,\dots,N$.
        \STATE Initialize $\w_1=\hat\w_0$ and $\Ll_1$ as a random TTm in site-$d$-mixed canonical format.\label{line:init}
        \FOR{$t=1,\dots,N$} 
        \STATE Compute $\hat{\mathbfcal{W}}^{(1)}_t,\hat{\mathbfcal{W}}^{(2)}_t,\dots,\hat{\mathbfcal{W}}^{(D)}_t$ with \eqref{eq:ALS}. \label{line:weights}
        \STATE Compute $\mathbfcal{L}^{(1)}_t,\mathbfcal{L}^{(2)}_t,\dots,\mathbfcal{L}^{(D)}_t$ with \eqref{eq:ALScov}. \label{line:cova}
        \STATE Recompute $\Ll_t$ with its orginal size with Algorithm \ref{alg:retruncate}. \label{line:qr}
        \STATE Compute $m_{*,t}$ with \eqref{eq:predpost} as depicted in Fig.\ \ref{fig:pred}(a). \label{line:predmean}
        \STATE Compute $\sigma^2_{*,t}$ with \eqref{eq:predpost} as depicted in Fig.\ \ref{fig:pred}(b). \label{line:predcova}
        \ENDFOR
    \end{algorithmic}
\end{algorithm}

\section{Implementation}
\label{sec:implementation}
In this section, we give a detailed description of the non-straight-forward TN operations to update the mean estimate $\hat\w_t$ and square root covriance factor $\Ll_t$ as described in Algorithm \ref{alg:onlineregression}.
The leading complexities of the mean and square root covriance factor update are given in Table \ref{tab:complexities}.


\begin{table}
    \caption{Computational complexities for one TT-core mean and covariance update. We denote the TT-ranks of $\K_t$ by $R_\K$.}
    \vspace{.2cm}
    \centering
    \begin{tabular}{cc}
    \toprule
         Term & Complexity \\ \midrule\midrule
         $\mathbf{G}_{d,t}^\top \hat\w_{t-1}$ & $\mathcal{O}(R_\w^4I)$  \\
         \;\;\;\;$\mathbf{G}_{d,t}^\top \K_t(y_t-\boldsymbol\phi_t^\top \hat\w_{t-1})$ \;\;\;\;&\;\;\;\; $\mathcal{O}(R_{\w}^2R_{\K}^2I)$ \;\;\;\;\;\\
         \eqref{eq:term1_eq}-\eqref{eq:secondtermeq} & $\mathcal{O}(R_\Ll^4IJ)$ \\
         \bottomrule
    \end{tabular}
    \label{tab:complexities}
\end{table}

\subsection{Updating $\hat\w_t$ in TN format}
In the following sections, we discuss the implementation of \eqref{eq:ALS} for the mean update (Algorithm \ref{alg:onlineregression}, line \ref{line:weights}), and we describe how the mean is initialized in TT format (Algorithm \ref{alg:onlineregression}, line \ref{line:init}).

\subsubsection{Implementation of $\mathbf{G}_{d,t}^\top(\hat\w_{t-1} + \mathbf{K}_t(y_t-\boldsymbol\phi_t^\top \hat\w_{t-1}))$}
To compute the TT representing the mean estimate $\hat\w_t$, we implement the ALS to solve \eqref{eq:ALS} (\mbox{Algorithm \ref{alg:onlineregression}, line \ref{line:weights}).}

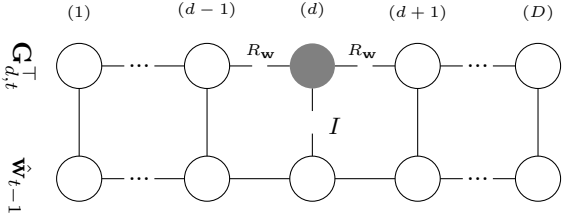
\begin{figure}
    \centering
    \begin{tikzpicture}[node distance = 2cm, auto,rotate=-90,transform shape]
\draw  (-1.2,2.7) ellipse (0.3 and 0.3);
\draw  (0.3,2.7) ellipse (0.3 and 0.3);
\draw  (-1.2,1.1) ellipse (0.3 and 0.3);
\draw  (0.3,1.1) ellipse (0.3 and 0.3);
\draw  (0.3,-0.3) ellipse (0.3 and 0.3);
\draw  (-1.2,-1.7) ellipse (0.3 and 0.3);
\draw  (0.3,-1.7) ellipse (0.3 and 0.3);
\draw  (-1.2,-3.4) ellipse (0.3 and 0.3);
\draw  (0.3,-3.4) ellipse (0.3 and 0.3);
\draw (-0.9,2.7) -- (0,2.7);
\draw (-0.9,1.1) -- (0,1.1);
\draw (-0.3,-0.3) -- (0,-0.3);
\draw (-0.9,-1.7) -- (0,-1.7);
\draw (-0.9,-3.4) -- (0,-3.4);
\draw (-1.2,1.7) -- (-1.2,1.4);
\draw (-1.2,0.8) -- (-1.2,0.5);
\draw (-1.2,-2) -- (-1.2,-2.4);
\draw (0.3,1.7) -- (0.3,1.4);
\draw (0.3,0.8) -- (0.3,0);
\draw (0.3,-0.6) -- (0.3,-1.4);
\draw (0.3,-2) -- (0.3,-2.4);
\node at (-1.2,-4.2) {$\mathbf{G}^\top_{d,t}$};
\node at (0.4,-4.2) {$\hat{\mathbf{w}}_{t-1}$};
\draw (-1.2,2.4) -- (-1.2,2.1);
\draw (0.3,2.4) -- (0.3,2.1);
\draw (-1.2,-3.1) -- (-1.2,-2.8);
\draw (0.3,-3.1) -- (0.3,-2.8);
\node[rotate=90] at (-1.2,1.9) {...};
\node[rotate=90] at (0.3,1.9) {...};
\node[rotate=90] at (-1.2,-2.6) {...};
\node [rotate=90] at (0.3,-2.6) {...};
\node [rotate=90] at (-1.9,-3.4) {\tiny$(1)$};
\node [rotate=90] at (-1.95,-1.7) {\tiny$(d-1)$};
\node [rotate=90] at (-1.95,-0.3) {\tiny$(d)$};
\node [rotate=90] at (-1.9,1.1) {\tiny$(d+1)$};
\node [rotate=90] at (-1.9,2.7) {\tiny$(D)$};
\fill [color=gray]  (-1.2,-0.3) ellipse (0.3 and 0.3);
\draw (-0.9,-0.3) -- (-0.6,-0.3);
\draw (-1.2,0) -- (-1.2,0.3);
\draw (-1.2,-0.6) -- (-1.2,-0.9);
\node[rotate=90] at (-1.4,-0.98) {\tiny$R_\mathbf{w}$};
\node[rotate=90] at (-1.4,0.37) {\tiny$R_\mathbf{w}$};
\node[rotate=90] at (-0.4,0) {\small$I$};
\draw (-1.2,-1.4) -- (-1.2,-1.1);
\end{tikzpicture}
    \caption{Visual depiction of computation of $\mathbf{G}_{d,t}^\top\hat\w_{t-1}$, resulting in three-way tensor of size $R_\w\times I \times R_\w$ (gray node). The indices are summed over from left to right, alternating between the vertical and horizontal ones.}
    \label{fig:updatemeanTN}
\end{figure}

The following example illustrates the update of one TT-core during the ALS.
\begin{exmp}\textbf{TT-core update with ALS}
    Take a $D=5$ dimensional weight vector in TT format with $M_1=M_2=M_3=M_4=M_5=10$ basis functions in each dimension, resulting in $10^5$ parameters and uniform TT-ranks of $R_2=R_3=R_4=4$.
    Say, we are currently updating the third TT-core $\mathbfcal{W}_t^{(3)}\in\mathbb{R}^{4\times10\times 4}$ using
    \begin{equation}
        \underbrace{\w_t^{(3)}}_{160\times1} = \underbrace{\mathbf{G}_{3,t}^\top}_{160\times 10^5}\left(\underbrace{\hat\w_{t-1}}_{10^5\times 1} + \underbrace{\mathbf{K}_t}_{10^5\times 1} \underbrace{(y_t-\boldsymbol\phi_t^\top \hat\w_{t-1})}_{1\times1}\right).
    \end{equation}
    We first multiply over the large dimension of $10^5$ in $\mathbf{G}_{3,t}^\top \hat\w_{t-1}$ and $\mathbf{G}_{3,t}^\top \K_t(y_t-\boldsymbol\phi_t^\top \hat\w_{t-1})$. 
    In TT format, this matrix-vector-multiplication is done core by core, thus avoiding 
    the explicit multiplication. 
    Finally, we sum two vectors of size $160$.
    \label{ex:mean}
\end{exmp}
Figure \ref{fig:updatemeanTN} illustrates the multiplication of $\mathbf{G}_{d,t}^\top \hat\w_{t-1}$ in TT format, resulting in a tensor of size $R_\w\times I\times R_\w$.
The multiplication of between $\mathbf{G}_{d,t}^\top$ and $\K_t(y_t-\boldsymbol\phi_t^\top \hat\w_{t-1})$ works in the same way after firstly computing $\boldsymbol\phi_t^\top \hat\w_{t-1}$ in TN format and secondly multiplying one arbitrary TT-core of $\K_t$ by the scalar $(y_t-\boldsymbol\phi_t^\top \hat\w_{t-1})$.

During the update of the $d$th TT-core, the TT is in site-$d$-mixed canonical format.
Before updating the next TT-core, either the $(d-1)$th or the $(d+1)$th, the site-$(d-1)$-mixed or site-$(d+1)$-mixed canonical format is computed.
Note that because of the recursive property, updating every TT-core once with a new measurement is usually sufficient for the residual of \eqref{eq:optproblem} to converge.

\subsubsection{Initialization of $\hat\w_0$ and $\w_1$}
For the first time step $t=1$ of Algorithm \ref{alg:onlineregression}, we choose a zero-mean assumption for the prior estimate $\hat\w_0$.
The following Lemma explains how this can be implemented in TT format.

\begin{lem}\textbf{Zero-mean prior in TT format \cite{Batselier2019extended}}
    Consider a vector with all entries equal to zero.
    In TT format, such a vector is given by a TT in site-$d$-mixed canonical format, where the $d$th TT-core contains only zeros. \label{lem:zmprior}
\end{lem}

In addition, Algorithm \ref{alg:onlineregression} requires an initial guess for $\w_1$ to compute $\mathbf{G}_{d,1}$ from all TT-cores of $\w_1$, except the $d$th.
For this, we set $\w_1=\hat\w_0$.

\subsection{Updating $\Ll_t$ in TT format}
\label{sec:implementcov}

To compute the TTm representing $\Ll_t$, we implement the ALS to solve \eqref{eq:ALScov} (Algorithm \ref{alg:onlineregression}, line \ref{line:cova}).
The following example illustrates the update of one TTm-core during the ALS.

\begin{exmp}\textbf{TTm-core update with ALS}\label{ex:covacoreupdate}
    Take a $D=5$ dimensional TTm representing $\Ll_t\in\mathbb{R}^{M\times M}$, where we are currently updating the third TTm-core. 
    We have $I=10$ and $J=10$, where the third TTm-core is augmented, and $R_\Ll=4$.
    We update $\mathbfcal{L}_t^{(3)}\in\mathbb{R}^{4\times10\times20\times 4}$ using
    \begin{equation}
    \begin{aligned}
        \underbrace{\mathbf{l}_t^{(3)}}_{3200\times1} =   \underbrace{\mathbf{H}_{d,t}^\top}_{3200\times2\cdot10^{10}}&\operatorname{vec}\left(\underbrace{\begin{bmatrix}1&0\end{bmatrix}}_{1\times 2}\otimes\underbrace{\Ll_{t-1}}_{10^5\times 10^5}\right)\\
        - \underbrace{\mathbf{H}_{d,t}^\top}_{3200\times2\cdot10^{10}} &\operatorname{vec}\left(\underbrace{\begin{bmatrix}1&0\end{bmatrix}}_{1\times 2} \otimes \underbrace{\K_t\boldsymbol\phi_t^\top\Ll_{t-1}}_{10^5\times 10^5}\right)\\
        + \underbrace{\mathbf{H}_{d,t}^\top}_{3200\times2\cdot10^{10}}&\operatorname{vec}\left(\underbrace{\begin{bmatrix}0&1\end{bmatrix}}_{1\times 2} \otimes \underbrace{\begin{bmatrix}1&\boldsymbol0_{M-1}\end{bmatrix}}_{1\times 10^5}\otimes\underbrace{\sigma_y\K_t}_{10^5\times 1}\right).
    \end{aligned}\label{eq:threeterms}
    \end{equation}
    We first multiply over the large dimension of $2\cdot10^{10}$ in TT format, then sum the three terms of size $3200\times 1$.
\end{exmp}
From Example \ref{ex:covacoreupdate}, it follows that the three terms of \eqref{eq:threeterms} need to be implemented.
We discuss them separately in the following sections.
We distinguish between the update of the augmented TTm-core from all other ones, which result in TTm-cores of size $R_\Ll\times I\times 2J \times R_\Ll$ and $R_\Ll\times I\times J\times R_\Ll$, respectively.
In the tensor diagrams (Fig.\ \ref{fig:term1}-\ref{fig:term3}), we depict the update for the augmented TTm-core.

Before diving in, recall from \eqref{eq:sitek_ttm} that $\mathbf{H}_{d,t}$ is computed from TTm-cores of $\Ll_t$, except the $d$th, where row and column indices are combined.
In the tensor diagrams, the indices are depicted not as combined because, in practice, they are generally summed over separately.
However, the vectorized format is necessary for writing down the equations in matrix form.

\subsubsection{Implementation of first term of \eqref{eq:ALScov}}
Fig.\ \ref{fig:term1} illustrates the computation of the augmented TTm-core in the first term of \eqref{eq:ALScov}, given by
\begin{equation}
    \mathbf{H}_{d,t}^\top \operatorname{vec}\left(\begin{bmatrix}1&0\end{bmatrix} \otimes \sigma_y\Ll_{t-1}\right).\label{eq:term1_eq}
\end{equation}
The column indices of $\Ll_{t-1}$ are indicated by the round edges that are connected to the row indices of $\mathbf{H}_{d,t}^\top$.
The edge containing $\mathbf{e}_1=[1\;\;0]$ is connected to the $d$th TTm core of $\Ll_t$ with a rank-1 connection, which corresponds to the Kronecker product in \eqref{eq:term1_eq}.
The summation over the vertical and curved indices has the leading computational complexity of $\mathcal{O}(R_\Ll^4IJ)$ per dimension.
When updating all TTm-cores except the augmented TTm-core, the additional index of size 2 is summed over resulting in a tensor of size $R_\Ll\times I\times J\times R_\Ll$.

\begin{figure}
    \centering
    \begin{tikzpicture}[node distance = 2cm, auto,rotate=-90,transform shape]
\draw  (-1.2,2.7) ellipse (0.3 and 0.3);
\draw  (0.3,2.7) ellipse (0.3 and 0.3);
\draw  (-1.2,1.1) ellipse (0.3 and 0.3);
\draw  (0.3,1.1) ellipse (0.3 and 0.3);
\draw  (0.3,-0.3) ellipse (0.3 and 0.3);
\draw  (-1.2,-1.7) ellipse (0.3 and 0.3);
\draw  (0.3,-1.7) ellipse (0.3 and 0.3);
\draw  (-1.2,-3.4) ellipse (0.3 and 0.3);
\draw  (0.3,-3.4) ellipse (0.3 and 0.3);
\draw (-0.9,2.7) -- (0,2.7);
\draw (-0.9,1.1) -- (0,1.1);
\draw (-0.3,-0.3) -- (0,-0.3);
\draw (-0.9,-1.7) -- (0,-1.7);
\draw (-0.9,-3.4) -- (0,-3.4);
\draw (-1.2,1.7) -- (-1.2,1.4);
\draw (-1.2,0.8) -- (-1.2,0.5);
\draw (-1.2,-2) -- (-1.2,-2.4);
\draw (0.3,1.7) -- (0.3,1.4);
\draw (0.3,0.8) -- (0.3,0);
\draw (0.3,-0.6) -- (0.3,-1.4);
\draw (0.3,-2) -- (0.3,-2.4);
\node at (-1.2,-4.2) {$\mathbf{H}^\top_{d,t}$};
\node at (0.4,-4.2) {$\mathbf{L}_{t-1}$};
\draw (-1.2,2.4) -- (-1.2,2.1);
\draw (0.3,2.4) -- (0.3,2.1);
\draw (-1.2,-3.1) -- (-1.2,-2.8);
\draw (0.3,-3.1) -- (0.3,-2.8);
\node[rotate=90] at (-1.2,1.9) {...};
\node[rotate=90] at (0.3,1.9) {...};
\node[rotate=90] at (-1.2,-2.6) {...};
\node [rotate=90] at (0.3,-2.6) {...};
\node [rotate=90] at (-1.9,-3.4) {\tiny$(1)$};
\node [rotate=90] at (-1.95,-1.7) {\tiny$(d-1)$};
\node [rotate=90] at (-1.95,-0.3) {\tiny$(d)$};
\node [rotate=90] at (-1.9,1.1) {\tiny$(d+1)$};
\node [rotate=90] at (-1.9,2.7) {\tiny$(D)$};
\fill [color=gray]  (-1.2,-0.3) ellipse (0.3 and 0.3);
\draw (-1.8,-0.3) -- (-1.5,-0.3);
\draw (-0.9,-0.3) -- (-0.6,-0.3);
\draw (-1.2,0) -- (-1.2,0.3);
\draw (-1.2,-0.6) -- (-1.2,-0.9);
\node[rotate=90] at (-1.4,-0.93) {\tiny$R_\mathbf{L}$};
\node[rotate=90] at (-1.68,-0.06) {\small$2J$};
\node[rotate=90] at (-1.4,0.35) {\tiny$R_\mathbf{L}$};
\node[rotate=90] at (-0.5,-0.1) {\small$I$};
\draw (-1.2,-1.4) -- (-1.2,-1.1);
\draw  plot[smooth, tension=.7] coordinates {(-1.5,-3.4) (-1.6,-3) (-0.4,-2.9) (0.7,-3) (0.6,-3.4)};
\draw  plot[smooth, tension=.7] coordinates {(-1.5,-1.7) (-1.6,-1.3) (-0.4,-1.2) (0.7,-1.3) (0.6,-1.7)};
\draw  plot[smooth, tension=.7] coordinates {(-1.5,1.1) (-1.6,1.5) (-0.4,1.6) (0.7,1.5) (0.6,1.1)};
\draw  plot[smooth, tension=.7] coordinates {(-1.5,2.7) (-1.6,3.1) (-0.4,3.2) (0.7,3.1) (0.6,2.7)};
\draw (0.6,-0.3) -- (0.9,-0.3);
\draw  (1.1,-0.9) ellipse (0.3 and 0.3);
\draw (0.56,-0.47) -- (0.83,-0.72);
\draw (1.4,-0.9) -- (1.7,-0.9);
\node[rotate=90] at (0.6,-0.7) {\tiny 1};
\node[rotate=90] at (1.6,-0.7) {\tiny 2};
\node[rotate=90] at (0.8,-0.1) {\small$J$};
\node[rotate=90] at (1.1,-0.9) {$\mathbf{e}_1$};
\end{tikzpicture}
    \caption{Visual depiction for computing the augmented TTm-core in \eqref{eq:term1_eq} resulting in a 4-way tensor of size $R_\Ll\times I\times 2J\times R_\Ll$ (gray node). The combined horizontal and curved indices are summed over and alternating with the horizontal indices. }
    \label{fig:term1}
\end{figure}

\subsubsection{Implementation of second term of \eqref{eq:ALScov}}
Fig.\ \ref{fig:term2} illustrates the computation of 
    \begin{equation}
        \mathbf{H}_{d,t}^\top \operatorname{vec}\left(\begin{bmatrix}1&0\end{bmatrix} \otimes\Ll_{t-1}\Ll_{t-1}^\top\boldsymbol{\phi}_t\Ss_t^{-1}\boldsymbol\phi_t^\top\Ll_{t-1}\right), \label{eq:term2_eq2}
    \end{equation}
which directly follows from the second term of \eqref{eq:ALScov}.
As shown, the row and column indices of $\mathbf{H}_{d,t}^\top$ are connected separately to the column and row indices of two TT matrices for $\Ll_{t-1}$, respectively.
Like in the previous term, the edge containing $\mathbf{e}_1=[1\;\;0]$ is connected to the augmented TTm-core of $\Ll_t$ with a rank-1 connection, which corresponds to the Kronecker product in \eqref{eq:term2_eq2}.
The leading computational complexity of $\mathcal{O}(R_\Ll^4IJ)$ per dimension comes from the summation over the vertical indices in the red or blue box indicated in the figure.
The most efficient order of doing the computations in Fig.\ \ref{fig:term2} was found with the visual tensor network software by \cite{evenbly2019tensortrace}.

\begin{figure}
    \centering
    \begin{tikzpicture}[node distance = 2cm, auto,rotate=-90,transform shape]
\draw  (-4.3,2.7) ellipse (0.3 and 0.3);
\draw  (-3.1,2.7) ellipse (0.3 and 0.3);
\draw  (-1.2,2.7) ellipse (0.3 and 0.3);
\draw  (0.3,2.7) ellipse (0.3 and 0.3);
\draw  (1.8,2.7) ellipse (0.3 and 0.3);
\draw  (3,2.7) ellipse (0.3 and 0.3);
\draw  (-4.3,1.1) ellipse (0.3 and 0.3);
\draw  (-3.1,1.1) ellipse (0.3 and 0.3);
\draw  (-1.2,1.1) ellipse (0.3 and 0.3);
\draw  (0.3,1.1) ellipse (0.3 and 0.3);
\draw  (1.8,1.1) ellipse (0.3 and 0.3);
\draw  (3,1.1) ellipse (0.3 and 0.3);
\draw  (-4.3,-0.3) ellipse (0.3 and 0.3);
\draw  (-3.1,-0.3) ellipse (0.3 and 0.3);
\draw  (0.3,-0.3) ellipse (0.3 and 0.3);
\draw  (1.8,-0.3) ellipse (0.3 and 0.3);
\draw  (3,-0.3) ellipse (0.3 and 0.3);
\draw  (-4.3,-1.7) ellipse (0.3 and 0.3);
\draw  (-3.1,-1.7) ellipse (0.3 and 0.3);
\draw  (-1.2,-1.7) ellipse (0.3 and 0.3);
\draw  (0.3,-1.7) ellipse (0.3 and 0.3);
\draw  (1.8,-1.7) ellipse (0.3 and 0.3);
\draw  (3,-1.7) ellipse (0.3 and 0.3);
\draw  (-4.3,-3.4) ellipse (0.3 and 0.3);
\draw  (-3.1,-3.4) ellipse (0.3 and 0.3);
\draw  (-1.2,-3.4) ellipse (0.3 and 0.3);
\draw  (0.3,-3.4) ellipse (0.3 and 0.3);
\draw  (1.8,-3.4) ellipse (0.3 and 0.3);
\draw  (3,-3.4) ellipse (0.3 and 0.3);
\draw (-4,2.7) -- (-3.4,2.7);
\draw (-2.8,2.7) -- (-1.5,2.7);
\draw (-0.9,2.7) -- (0,2.7);
\draw (0.6,2.7) -- (1.5,2.7);
\draw (2.1,2.7) -- (2.7,2.7);
\draw (-4,1.1) -- (-3.4,1.1);
\draw (-2.8,1.1) -- (-1.5,1.1);
\draw (-0.9,1.1) -- (0,1.1);
\draw (0.6,1.1) -- (1.5,1.1);
\draw (2.1,1.1) -- (2.7,1.1);
\draw (-4,-0.3) -- (-3.4,-0.3);
\draw (-2.8,-0.3) -- (-2.5,-0.3);
\draw (-0.3,-0.3) -- (0,-0.3);
\draw (0.6,-0.3) -- (1.5,-0.3);
\draw (2.1,-0.3) -- (2.7,-0.3);
\draw (-4,-1.7) -- (-3.4,-1.7);
\draw (-2.8,-1.7) -- (-1.5,-1.7);
\draw (-0.9,-1.7) -- (0,-1.7);
\draw (0.6,-1.7) -- (1.5,-1.7);
\draw (2.1,-1.7) -- (2.7,-1.7);
\draw (-4,-3.4) -- (-3.4,-3.4);
\draw (-2.8,-3.4) -- (-1.5,-3.4);
\draw (-0.9,-3.4) -- (0,-3.4);
\draw (0.6,-3.4) -- (1.5,-3.4);
\draw (2.1,-3.4) -- (2.7,-3.4);
\draw (-3.1,1.7) -- (-3.1,1.4);
\draw (-3.1,0.8) -- (-3.1,0);
\draw (-3.1,-0.6) -- (-3.1,-1.4);
\draw (-1.2,-1.1) -- (-1.2,-1.4);
\draw (-3.1,-2) -- (-3.1,-2.4);
\draw (-1.2,1.7) -- (-1.2,1.4);
\draw (-1.2,0.8) -- (-1.2,0.5);
\draw (-1.2,-2) -- (-1.2,-2.4);
\draw (0.3,1.7) -- (0.3,1.4);
\draw (0.3,0.8) -- (0.3,0);
\draw (0.3,-0.6) -- (0.3,-1.4);
\draw (0.3,-2) -- (0.3,-2.4);
\draw (1.8,1.7) -- (1.8,1.4);
\draw (1.8,0.8) -- (1.8,0);
\draw (1.8,-0.6) -- (1.8,-1.4);
\draw (1.8,-2) -- (1.8,-2.4);
\node  at (-4.2,-4.2) {$\boldsymbol{\phi}_t^\top$};
\node at (-3.1,-4.2) {$\mathbf{L}_{t-1}$};
\node at (-1.2,-4.4) {$\mathbf{H}^\top_{d,t}$};
\node at (0.4,-4.2) {$\mathbf{L}_{t-1}$};
\node at (1.9,-4.2) {$\mathbf{L}_{t-1}^\top$};
\node at (3,-4.2) {$\boldsymbol{\phi}_t$};
\draw (-3.1,2.4) -- (-3.1,2.1);
\draw (-1.2,2.4) -- (-1.2,2.1);
\draw (0.3,2.4) -- (0.3,2.1);
\draw (1.8,2.4) -- (1.8,2.1);
\draw (-3.1,-3.1) -- (-3.1,-2.8);
\draw (-1.2,-3.1) -- (-1.2,-2.8);
\draw (0.3,-3.1) -- (0.3,-2.8);
\draw (1.8,-3.1) -- (1.8,-2.8);
\node[rotate=90] at (-3.1,1.9) {...};
\node[rotate=90] at (-1.2,1.9) {...};
\node[rotate=90] at (0.3,1.9) {...};
\node[rotate=90] at (1.8,1.9) {...};
\node [rotate=90] at (-3.1,-2.6) {...};
\node[rotate=90] at (-1.2,-2.6) {...};
\node [rotate=90] at (0.3,-2.6) {...};
\node[rotate=90] at (1.8,-2.6) {...};
\node [rotate=90] at (-4.8,-3.4) {\tiny$(1)$};
\node [rotate=90] at (-4.8,-1.7) {\tiny$(d-1)$};
\node [rotate=90] at (-4.8,-0.3) {\tiny$(d)$};
\node [rotate=90] at (-4.8,1.1) {\tiny$(d+1)$};
\node [rotate=90] at (-4.8,2.7) {\tiny$(D)$};

\fill [color=gray]  (-1.2,-0.3) ellipse (0.3 and 0.3);
\draw (-1.8,-0.3) -- (-1.5,-0.3);
\draw (-0.9,-0.3) -- (-0.6,-0.3);
\draw (-1.2,0) -- (-1.2,0.3);
\draw (-1.2,-0.6) -- (-1.2,-0.9);
\node[rotate=90] at (-1.4,-0.8) {\tiny$R_\mathbf{L}$};
\node[rotate=90] at (-1.68,-0.07) {\small$2J$};
\node[rotate=90] at (-1.4,0.2) {\tiny$R_\mathbf{L}$};
\node[rotate=90] at (-0.4,-0.15) {\small$I$};
\draw  (-0.5,-4.8) ellipse (0.3 and 0.3);
\draw  plot[smooth, tension=.7] coordinates {(-0.8,-4.8) (-4.2,-4.5) (-4.6,-3.4)};
\draw  plot[smooth, tension=.7] coordinates {(-0.2,-4.8) (2.9,-4.6) (3.3,-3.4)};
\node  at (-0.5,-4.8) {\small$\mathbf{S}^{-1}_t$};
\draw  [color=red](-5,-3.8) rectangle (3.5,-1.2);
\draw  [color=blue](-5,0.6) rectangle (3.5,3.1);
\draw  [color=green](-5,-1.1) rectangle (-1.9,0.2);
\draw  [color=yellow](-0.2,-0.8) rectangle (3.5,0.2);

\node [rotate=90] at (-2.6,-0.1) {\small $J$};
\draw  (-2.5,-0.8) ellipse (0.3 and 0.3);
\draw (-2.89,-0.51) -- (-2.75,-0.65);
\draw (-2.2,-0.8) -- (-2,-0.8);
\node [rotate=90] at (-2.75,-0.5) {\tiny 1};
\node [rotate=90] at (-2.1,-0.7) {\tiny2};
\node [rotate=90] at (-2.5,-0.8) {$\mathbf{e}_1$};
\end{tikzpicture}
    \caption{Visual depiction for computing \eqref{eq:term2_eq2} resulting in a 4-way tensor of size $R_\Ll\times I\times 2J\times R_\Ll$ (gray node). First, the indices in the red and blue boxes are summed over, then the indices between the red, yellow, and blue boxes, and finally, the ones between the red, green, and blue boxes.}
    \label{fig:term2}
\end{figure}

\subsubsection{Implementation of third term of \eqref{eq:ALScov}}
Fig.\ \ref{fig:term3} illustrates the computation of 
\begin{equation}
   \mathbf{H}_{d,t}^\top \operatorname{vec}\left(\begin{bmatrix}0& 1\end{bmatrix}\otimes\begin{bmatrix}1&\boldsymbol0_{M-1}\end{bmatrix}\otimes\sigma_y\Ll_t\Ll_t^\top\boldsymbol{\phi}_t\Ss_t^{-1}\right),
   \label{eq:secondtermeq}
\end{equation}
which directly follows from the third term of \eqref{eq:ALScov}.
The row of nodes each filled with $\mathbf{e}_1=[1\;\;\boldsymbol{0}_{J-1}]$ corresponds to $[1\;\;\mathbf{0}_{M-1}]$ from \eqref{eq:secondtermeq} and their rank-1 connections to the nodes above is the second Kronecker product in \eqref{eq:secondtermeq}, which is done dimension-wise in TT format.
The node with $\mathbf{e}_2$ corresponds to $[0\;\;1]$ from \eqref{eq:secondtermeq} and its rank-1 connection is the first Kronecker product in \eqref{eq:secondtermeq}. 
The summation over the vertical indices is the leading computational complexity of $\mathcal{O}(R_\Ll^4IJ)$ per dimension.

\begin{figure}
    \centering
    \begin{tikzpicture}[node distance = 2cm, auto,rotate=-90,transform shape]
\draw  (-1.2,2.7) ellipse (0.3 and 0.3);
\draw  (0.3,2.7) ellipse (0.3 and 0.3);
\draw  (1.8,2.7) ellipse (0.3 and 0.3);
\draw  (3,2.7) ellipse (0.3 and 0.3);
\draw  (-1.2,1.1) ellipse (0.3 and 0.3);
\draw  (0.3,1.1) ellipse (0.3 and 0.3);
\draw  (1.8,1.1) ellipse (0.3 and 0.3);
\draw  (3,1.1) ellipse (0.3 and 0.3);
\draw  (0.3,-0.3) ellipse (0.3 and 0.3);
\draw  (1.8,-0.3) ellipse (0.3 and 0.3);
\draw  (3,-0.3) ellipse (0.3 and 0.3);
\draw  (-1.2,-1.7) ellipse (0.3 and 0.3);
\draw  (0.3,-1.7) ellipse (0.3 and 0.3);
\draw  (1.8,-1.7) ellipse (0.3 and 0.3);
\draw  (3,-1.7) ellipse (0.3 and 0.3);
\draw  (-1.2,-3.4) ellipse (0.3 and 0.3);
\draw  (0.3,-3.4) ellipse (0.3 and 0.3);
\draw  (1.8,-3.4) ellipse (0.3 and 0.3);
\draw  (3,-3.4) ellipse (0.3 and 0.3);
\draw (-0.9,2.7) -- (0,2.7);
\draw (0.6,2.7) -- (1.5,2.7);
\draw (2.1,2.7) -- (2.7,2.7);
\draw (-0.9,1.1) -- (0,1.1);
\draw (0.6,1.1) -- (1.5,1.1);
\draw (2.1,1.1) -- (2.7,1.1);
\draw (-0.3,-0.3) -- (0,-0.3);
\draw (0.6,-0.3) -- (1.5,-0.3);
\draw (2.1,-0.3) -- (2.7,-0.3);
\draw (-0.9,-1.7) -- (0,-1.7);
\draw (0.6,-1.7) -- (1.5,-1.7);
\draw (2.1,-1.7) -- (2.7,-1.7);
\draw (-0.9,-3.4) -- (0,-3.4);
\draw (0.6,-3.4) -- (1.5,-3.4);
\draw (2.1,-3.4) -- (2.7,-3.4);
\draw (-1.2,1.7) -- (-1.2,1.4);
\draw (-1.2,0.8) -- (-1.2,0.5);
\draw (-1.2,-2) -- (-1.2,-2.4);
\draw (0.3,1.7) -- (0.3,1.4);
\draw (0.3,0.8) -- (0.3,0);
\draw (0.3,-0.6) -- (0.3,-1.4);
\draw (0.3,-2) -- (0.3,-2.4);
\draw (1.8,1.7) -- (1.8,1.4);
\draw (1.8,0.8) -- (1.8,0);
\draw (1.8,-0.6) -- (1.8,-1.4);
\draw (1.8,-2) -- (1.8,-2.4);
\node at (-1.2,-4.2) {$\mathbf{H}^\top_{d,t}$};
\node at (0.4,-4.2) {$\mathbf{L}_{t-1}$};
\node at (1.9,-4.2) {$\mathbf{L}_{t-1}^\top$};
\node at (3,-4.2) {$\boldsymbol{\phi}_t$};
\draw (-1.2,2.4) -- (-1.2,2.1);
\draw (0.3,2.4) -- (0.3,2.1);
\draw (1.8,2.4) -- (1.8,2.1);
\draw (-1.2,-3.1) -- (-1.2,-2.8);
\draw (0.3,-3.1) -- (0.3,-2.8);
\draw (1.8,-3.1) -- (1.8,-2.8);
\node[rotate=90] at (-1.2,1.9) {...};
\node[rotate=90] at (0.3,1.9) {...};
\node[rotate=90] at (1.8,1.9) {...};
\node[rotate=90] at (-1.2,-2.6) {...};
\node [rotate=90] at (0.3,-2.6) {...};
\node[rotate=90] at (1.8,-2.6) {...};
\node [rotate=90] at (-2.05,-3.4) {\tiny$(1)$};
\node [rotate=90] at (-2.1,-1.7) {\tiny$(d-1)$};
\node [rotate=90] at (-2.1,-0.3) {\tiny$(d)$};
\node [rotate=90] at (-2.05,1.1) {\tiny$(d+1)$};
\node [rotate=90] at (-2.05,2.7) {\tiny$(D)$};
\fill [color=gray]  (-1.2,-0.3) ellipse (0.3 and 0.3);
\draw (-1.8,-0.3) -- (-1.5,-0.3);
\draw (-0.9,-0.3) -- (-0.6,-0.3);
\draw (-1.2,0) -- (-1.2,0.3);
\draw (-1.2,-0.6) -- (-1.2,-0.9);
\node[rotate=90] at (-1.4,-0.8) {\tiny$R_\mathbf{L}$};
\node[rotate=90] at (-1.8,0) {\small$2J$};
\node[rotate=90] at (-1.4,0.3) {\tiny$R_\mathbf{L}$};
\node[rotate=90] at (-0.5,-0.1) {\small$I$};
\draw  (4.13,-4.19) ellipse (0.6 and 0.3);
\node  at (4.13,-4.19) {\small$\sigma_y\mathbf{S}^{-1}_t$};
\draw  (3.9,2.7) ellipse (0.3 and 0.3);
\draw  (3.9,1.1) ellipse (0.3 and 0.3);
\draw  (3.9,-0.3) ellipse (0.3 and 0.3);
\draw  (3.9,-3.4) ellipse (0.3 and 0.3);
\draw  (3.9,-1.7) ellipse (0.3 and 0.3);
\node [rotate=90] at (3.9,2.7) {$\mathbf{e}_1$};
\node [rotate=90] at (3.9,1.1) {$\mathbf{e}_1$};
\node [rotate=90] at (3.9,-0.3) {$\mathbf{e}_1$};
\node [rotate=90] at (3.9,-3.4) {$\mathbf{e}_1$};
\node [rotate=90] at (3.9,-1.7) {$\mathbf{e}_1$};
\draw (-1.2,-1.4) -- (-1.2,-1.1);
\draw (3.3,-0.3) -- (3.6,-0.3);
\node [rotate=90] at (3.45,-0.13) {\tiny 1};
\draw (3.2,-3.6) -- (3.53,-4.19);
\node [rotate=90] at (3.44,-3.8) {\tiny 1};
\node [rotate=90] at (4.4,-0.1) {\small$J$};
\node [rotate=90] at (1.1,3.4) {\small $J$};
\node[rotate=90] at (1.1,1.8) {\small$J$};
\node[rotate=90] at (1.1,-2.7) {\small$J$};
\node[rotate=90] at (1.1,-1) {\small$J$};
\draw  (4.5,-0.8) ellipse (0.3 and 0.3);
\draw (4.02,-0.58) -- (4.2,-0.8);
\draw (4.8,-0.8) -- (5,-0.8);
\node[rotate=90] at (4,-0.8) {\tiny 1};
\node[rotate=90] at (4.92,-0.64) {\tiny 2};
\node [rotate=90] at (4.5,-0.8) {$\mathbf{e}_2$};
\draw (3.3,-3.4) -- (3.6,-3.4);
\draw (3.3,-1.7) -- (3.6,-1.7);
\draw (3.3,1.1) -- (3.6,1.1);
\draw (3.3,2.7) -- (3.6,2.7);
\node [rotate=90] at (3.45,-3.23) {\tiny1};
\node [rotate=90] at (3.45,-1.53) {\tiny1};
\node [rotate=90] at (3.45,1.27) {\tiny1};
\node [rotate=90] at (3.45,2.87) {\tiny1};
\draw  plot[smooth, tension=.7] coordinates {(-1.5,-3.4) (-0.9,-2.9) (3.6,-2.9) (4.2,-3.4)};
\draw  plot[smooth, tension=.7] coordinates {(-1.5,-1.7) (-0.9,-1.2) (3.6,-1.2) (4.2,-1.7)};
\draw  plot[smooth, tension=.7] coordinates {(-1.5,1.1) (-0.9,1.6) (3.6,1.6) (4.2,1.1)};
\draw (4.2,-0.3) -- (4.6,-0.3);
\draw  plot[smooth, tension=.7] coordinates {(-1.5,2.7) (-0.9,3.2) (3.6,3.2) (4.2,2.7)};
\end{tikzpicture}
    \caption{Visual depiction for computing \eqref{eq:secondtermeq}, resulting in a 4-way tensor of size $R_\Ll\times I\times 2J\times R_\Ll$ (gray node). 
    The indices are summed over from left to right by alternating between the vertical and horizontal ones.}
    \label{fig:term3}
\end{figure}

\subsubsection{SVD-based QR step in TTm format}
\label{sec: QRstep_square}
When computing \eqref{eq:ALScov}, we double the number of columns of $\Ll_t$ compared to $\Ll_{t-1}$.
For the next time step, however, we need to transform $\Ll_t$ back to its original size (Algorithm \ref{alg:onlineregression}, line \ref{line:qr}), otherwise its column size will grow with the iterations and slow down the algorithm.
The QR step, as in \eqref{eq:augL}, computes a full QR decomposition of $\Ll_t$, which cannot be done in TT format.
Instead, we compute a thin SVD in TTm format (Definition \ref{def:svdthin}) of $\Ll_t$ transformed into a tall TTm (here also denoted by $\Ll_t$) with all row indices of size $IJ$, except the $d$th which is of size $I$, and the $d$th column index of size $2J$.
The $J$-truncated SVD of $\Ll_t$ is then given by
\begin{equation}
    \underbrace{\Ll_t}_{MJ^{D-1}\times2J}\approx\underbrace{\mathbf{U}_t\mathbf{S}_t}_{MJ^{D-1}\times J}\underbrace{\mathbf{V}_t^\top}_{J\times2J}, \label{eq:qrsteptall}
\end{equation}
where $\mathbf{U}_t\mathbf{S}_t$ is the new $\Ll_t$ and $\mathbf{V}_t^\top$ can be discarded because of \eqref{eq:covfromQR}.
In practice, we compute an SVD of the augmented TTm-core and truncate it back to the size of $R_\Ll\times I\times J\times R_\Ll$.

There is a way to make \eqref{eq:qrsteptall} exact.
This is possible if the augmented TTm-core is of size $R_\Ll\times I\times 2J R_\Ll^2\times R_\Ll$.
In this case, the SVD computed of the augmented TTm-core results in a square $\mathbf{U}$-factor.
Since the number of columns is doubled every measurement update, the QR step can be skipped $p$ times until $2^p=2R_\Ll^2$.
Choosing smaller values for $p$ reduces computational complexity at the cost of accuracy.

The SVD-based QR step is described in Algorithm \ref{alg:retruncate}.
The SVD of the reshaped and permuted augmented TTm-core is truncated for $2^p<2R_\Ll^2$ and exact for $2^p\ge 2R_\Ll^2$.

\begin{algorithm}
\caption{SVD-based QR step of covariance update}\label{alg:retruncate}
    \begin{algorithmic}[1]\onehalfspacing
    \REQUIRE TTm $\Ll_t$ in site-$d$-mixed canonical format with $\mathbfcal{L}^{(d)}\in\mathbb{R}^{R_\Ll\times I\times 2^pJ \times R_\Ll}$.
    \ENSURE TTm $\Ll_t$ with $\mathbfcal{L}^{(d)}\in\mathbb{R}^{R_\Ll\times I\times 2^{p-1}J\times R_\Ll}.$ 
        \STATE $\Ll^{(d)}\gets$ Reshape / permute $\mathbfcal{L}^{(d)}$ into matrix of size $R_\Ll IR_\Ll\times 2^{p+1}J$.
        \STATE Compute thin SVD($\Ll^{(d)})= \mathbf{U}^{(d)}\mathbf{S}^{(d)}(\mathbf{V}^{(d)})^\top$.
        \STATE $\mathbfcal{L}^{(d)}\gets$ Reshape / permute first $2^{p-1}J$ columns of $\mathbf{U}^{(d)}\mathbf{S}^{(d)}$ of size $R_\Ll IR_\Ll\times 2^{p-1}J$ into tensor of size $R_\Ll\times I\times 2^{p-1}J\times R_\Ll$.
    \end{algorithmic}
\end{algorithm}

\subsubsection{Initialization of $\Ll_0$ and $\Ll_1$}
At time $t=1$, Algorithm \ref{alg:onlineregression} requires the prior square root covariance factor $\Ll_0$ in TTm format.
We are considering product kernels that have priors in Kronecker format.
The following Lemma describes how these types of priors can be transformed into a TTm for $\Ll_0$.

\begin{lem}
\textbf{(Prior covariance with Kronecker structure into TTm, follows from \cite[p.708]{golub2013matrix})}
Given a prior covariance $\Pp_0=\Pp_0^{(1)}\otimes \Pp_0^{(2)} \otimes \cdots \otimes \Pp_0^{(D)}$, the prior square root covariance in TTm format is given by a TTm with all ranks equal to 1, where the cores are given by $\Ll_0^{(1)},\Ll_0^{(2)},\dots,\Ll_0^{(D)}$, each reshaped into a 4-way tensor of size $1\times I\times J\times 1$.
\label{lem:srcovM}
\end{lem}

In addition, Algorithm \ref{alg:onlineregression} requires an initial guess in TTm format for $\Ll_1\in\mathbb{R}^{M\times 2M}$.
We cannot set $\Ll_1 = \Ll_0$ since the prior has TTm-ranks equal to one, and we may want higher TTm-ranks for $\Ll_t$. 
This is because the choice of the TTm-ranks of $\Ll_1$ determines the rank manifold on which the TTm-cores will be optimized.
We initialize the TTm-cores as random samples from a zero-mean Gaussian distribution and transform the TTm into site-$d$-mixed canonical format, where $d$ is the augmented TTm-core.
\section{Experiments}
In this section, we show how our method works in practice by performing online GP regression on synthetic and real-life data sets.
We evaluate our predictions based on the root mean square error (RMSE) for the accuracy of the mean and negative log-likelihood (NLL) for the uncertainty estimation.
The metrics after $t$ measurement updates are defined as
\begin{equation}
    \begin{aligned}
        (\mathrm{RMSE})_t &= \sqrt{\sum_{i=1}^{N_*}\frac{(m_{*,t,i}-y_{*,i})^2}{N_*}} \;\;\; \mathrm{and}\\
        (\mathrm{NLL})_t &= 0.5\sum_{i=1}^{N_*} \log (2\pi\sigma_{*,t,i}^2) + \frac{(m_{*,t,i}-y_{*,i})^2}{\sigma_{*,t,i}^2},
    \end{aligned}
\end{equation}
where $y_{*,i}$ is the $i$th measurement from the test set, $m_{*,t,i}$ and $\sigma_{*,t,i}$ are the predictive mean and variance for the $i$th test point, and $N_*$ is the number of test points.

First, we show the equivalence of the full-rank TNSRKF and the conventional Kalman filter.
Then we show in a synthetic experiment how the choice of $R_\w$ and $R_\Ll$ impact the accuracy of the approximation.
Finally, we compare our method to the TNKF on a benchmark data set for nonlinear system identification.

All experiments were performed on an 11th Gen Intel(R) Core(TM) i7 processor running at 3.00 GHz with 16 GB RAM.
For reproducibility of the method and the experiments, the code written in Julia programming language is freely available at \mbox{\url{https://github.com/clarazen/TNSRKF}}.

\subsection{Equivalence of full-rank TNSRKF and Kalman filter}
\begin{table}[]
    \caption{RMSE and NLL at time $t=N$ for the full-rank setting and different choices of $p$ in comparison to the conventional Kalman filter (KF).}
    \centering 
    \begin{tabular}{cccccc}
    \toprule    
         Method&\multicolumn{3}{c}{Setting}& (RMSE)$_N$ & (NLL)$_N$ \\ \midrule\midrule
         KF &&-&& 0.07873 & -106.864\\ 
         \;\;TNSRKF\;\; &$R_\w$& $R_\Ll$ &$p$ & & \\
         & 4 & 16 & 8 & 0.07873 & -106.864 \\
         & 4 & 16 & 4 & 0.07879 & -108.338 \\
         & 4 & 16 & 2 & 0.07444 & -157.716\\
         & 4 & 16 & 1 & 0.06765 & -166.178 \\
         \bottomrule
    \end{tabular}
    \label{tab:comparetoKF}
\end{table}
In the first experiment, we show in which case our method is equivalent to the measurement update of the conventional Kalman filter.
We generate $D=3$ dimensional synthetic data sampled from a reduced-rank GP by \cite{HilbertGP} with a squared exponential kernel and use $I=4$ basis functions per dimension, such that $\Pp_t\in\mathbb{R}^{64\times64}$.
The input data lies in a cuboid given by $[-1\;\;1]\times[-1\;\;1]\times[-1\;\;1]$ and $N,N_*=100$

Table \ref{tab:comparetoKF} shows the RMSE and NLL for test data at time $t=N$ for different choices of $p$.
The TNSRKF is equivalent to the Kalman filter when both $R_\w$ and $R_\Ll$ are full-rank.
In addition, $p$ must be chosen, such that the QR step, discussed in Section \ref{sec: QRstep_square}, is exact.
For settings with lower values for $p$, the method trade-in accuracy.

In the following sections, we look at scenarios where the Kalman filter can no longer be computed on a conventional laptop because both storage and computational time become unfeasible.

\subsection{Influence of the ranks on the approximation}
The choice of the TT- and TTm-ranks is not obvious and can be intricate.
However, the computational budget often determines how high the ranks can be chosen.
In this experiment, we use our method to make online GP predictions on synthetic data while varying the TTm-ranks of $\Ll_t$, as well as the TT-ranks of $\w_t$.

We consider the Volterra kernel, a popular choice for nonlinear system identification.
It is known that the truncated Volterra series suffers from the curse of dimensionality, which was lifted in a TN setting by \cite{batselier2017VoleterrALS}.
With the notation of this paper, the basis functions $\boldsymbol{\phi}$ of parametric model \eqref{eq:parammodel} are a combination of monomials computed from the input sequence of the given problem.
We generate synthetic training and testing data as described in \cite{batselier2021enforcing}, where $D=7$ and $I=4$ such that the number of parameters is $4^7=16\thinspace384$.
We set the SNR to 60, so there is relatively little noise.

\begin{figure}
    \centering
    \includegraphics[width=\columnwidth]{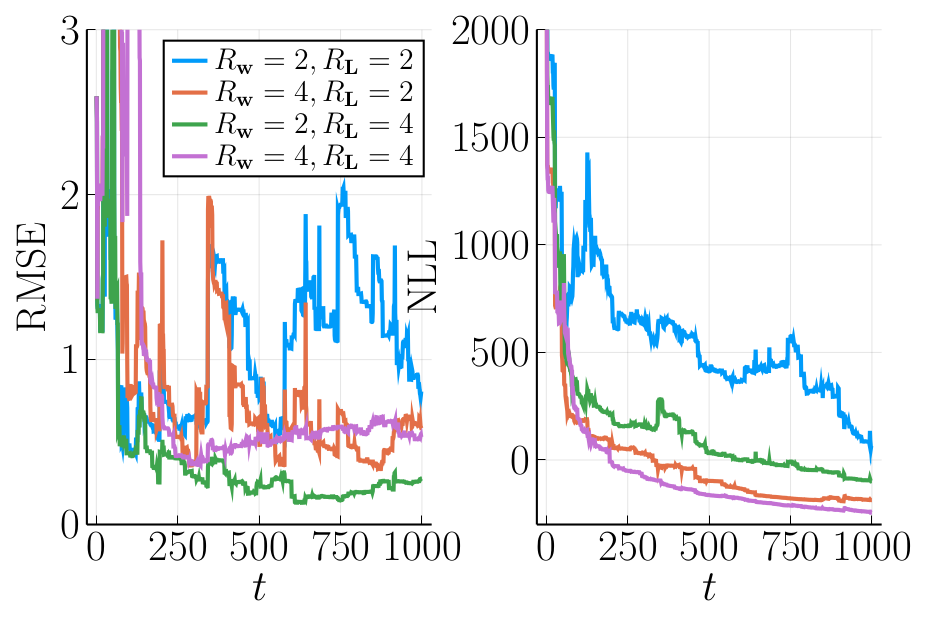}
    \caption{RMSE and NLL over time iterations for different combinations of $R_\w$ and $R_\Ll$.}
    \label{fig:rankinfluence}
\end{figure}

Fig.\ \ref{fig:rankinfluence} shows the RMSE and NLL on the testing data for $R_\w=2,4$ and $R_\Ll=2,4$ over time iterations of the TNSRKF.
At $t=N$, the RMSE is lower for $R_\w=2$ and $R_\Ll=4$ than for $R_\w=4$ and $R_\Ll=4$.
Thus it seems that a lower value for the mean estimate represents the data better.
The NLL is the lowest for $R_\w=4$ and $R_\Ll=4$, which is close to the NLL for $R_\w=4$ and $R_\Ll=2$.
Note that the NLL for the same $R_\Ll$ is different for the two settings of $R_\w$, because the NLL also depends on the difference between predicted and actual measurements, thus on the accuracy of $\hat\w_t$.

This experiment showed that the choice of $R_\w$ and $R_\Ll$ influence the performance of the TNSRKF.
Since higher values for the ranks also increase the computational complexity, the computational budget will determine the higher limit for the ranks.
In addition, an assumption with lower ranks may be fitting the data better in some cases.

\subsection{Comparison to TNKF for cascaded tanks benchmark data set}
In this experiment, we compare our method to the TNKF on a nonlinear benchmark for system identification, the cascaded tanks data set. 
A detailed description can be found in \cite{schoukens2017three}.
The training and testing data set consists of 1024 data points.
To train our GP model, we choose lagged inputs and outputs as input to our GP, as described in \cite{karagoz2020nonlinear}, resulting in an input of dimensionality $D=14$.
We use a squared exponential kernel, which hyperparameters we optimize with the Gaussian process toolbox by \cite{Rasmussen2006}, and we choose $I=4$, such that the model has $M=4^{14}=268\thinspace435\thinspace456$ parameters.

\begin{figure}
    \centering
    \includegraphics[width=\columnwidth]{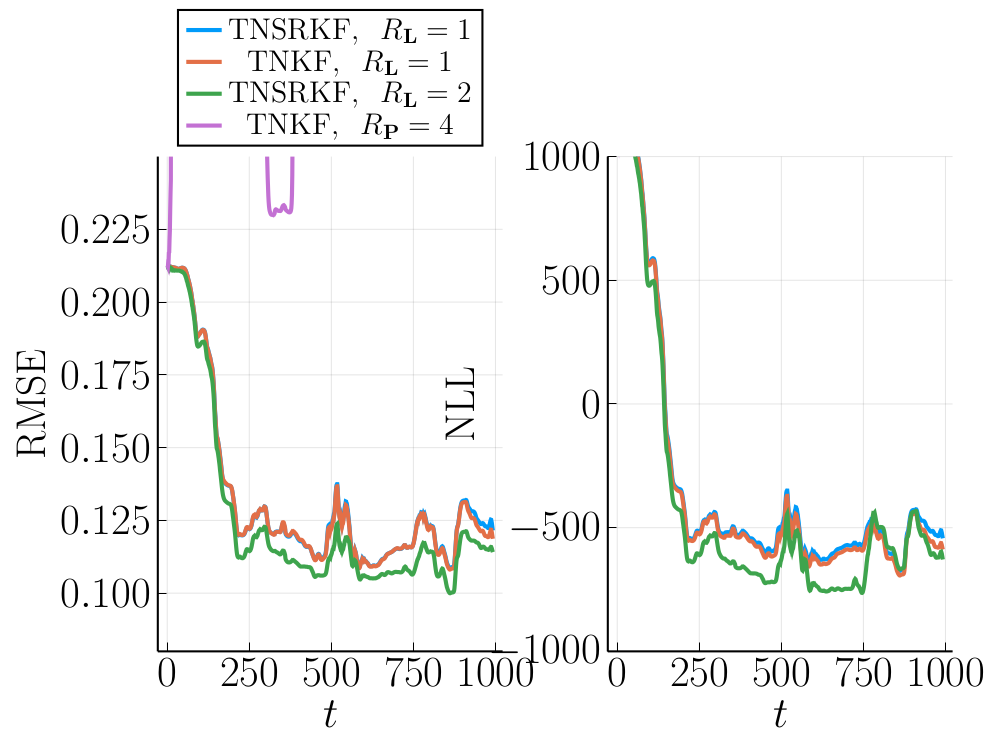}
    \caption{RMSE and NLL over iterations for TNKF and TNSRKF for $R_\Ll=R_\Pp=1$ and $R_\Ll=2$, $R_\Pp=R_\Ll\cdot R_\Ll=4$. The green and blue lines mostly overlap because both methods perform similarly for $R_\Ll=R_\Pp=1$. Also, the violet curve leaves the plot window because the TNKF diverges.}
    \label{fig:watertanks_comparison1}
\end{figure}

\begin{figure}
    \centering
    \includegraphics[width=\columnwidth]{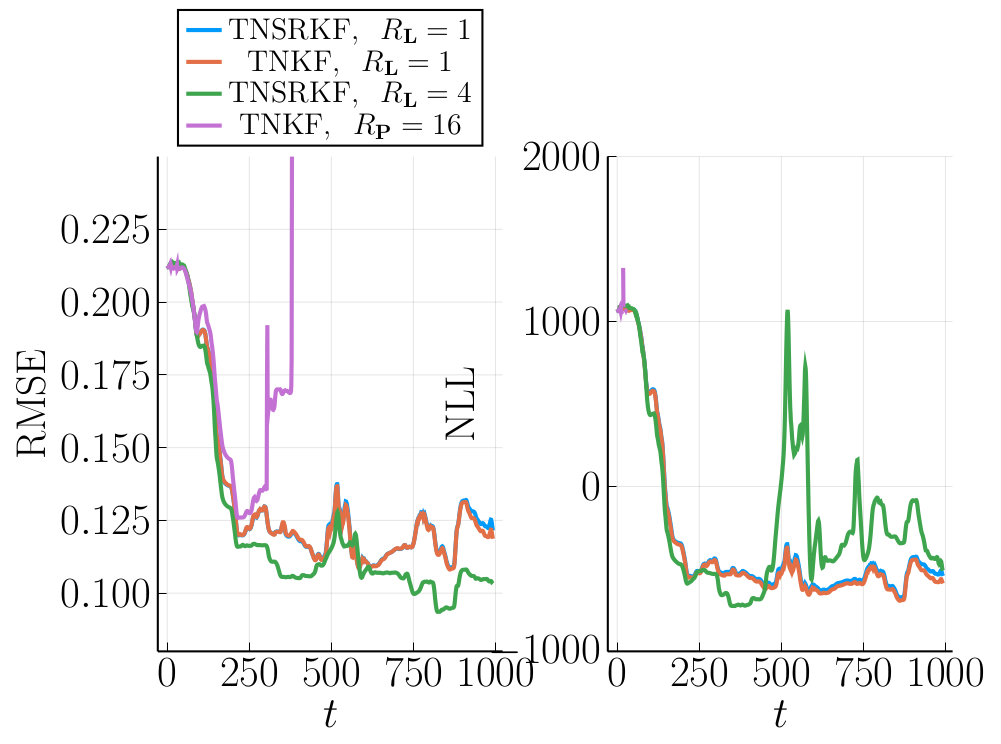}
    \caption{RMSE and NLL over iterations for TNKF and TNSRKF for $R_\Ll=R_\Pp=1$ and $R_\Ll=4$, $R_\Pp=R_\Ll\cdot R_\Ll=16$. The green and blue lines mostly overlap because both methods perform similarly for $R_\Ll=R_\Pp=1$. Also, the violet curve leaves the plot window because the TNKF diverges.}
    \label{fig:watertanks_comparison2}
\end{figure}

For the comparison to the TNKF, we choose the TT-ranks for the mean to be $R_2=R_{14}=4$, $R_3=\cdots R_{13}=10$, and we vary $R_\Ll$ and the TTm-ranks of the covariance matrix for the TNKF denoted by $R_\Pp$.
Fig.\ \ref{fig:watertanks_comparison1} and \ref{fig:watertanks_comparison2} show the RMSE and NLL over the time iterations of the respective filter. 
When $R_\Ll=1$ and $R_\Pp=1$, both methods perform almost the same, as visualized by the overlapping green and blue lines.
When $R_\Ll^2=R_\Pp=4$, our method improves both prediction accuracy and uncertainty estimation compared to the $R_\Ll=1$.
On the contrary, the TNKF diverges and leaves the plotted figure area because the covariance matrix loses positive definiteness.
When $R_\Ll^2=R_\Pp=16$, the TNKF shows a similar behavior, while the TNSRKF results in lower RMSEs but mostly higher NLL values.
This setting shows that higher values for $R_\Ll$ are not always beneficial for the uncertainty estimation.

\begin{figure}
   \centering
   \includegraphics[width=\linewidth]{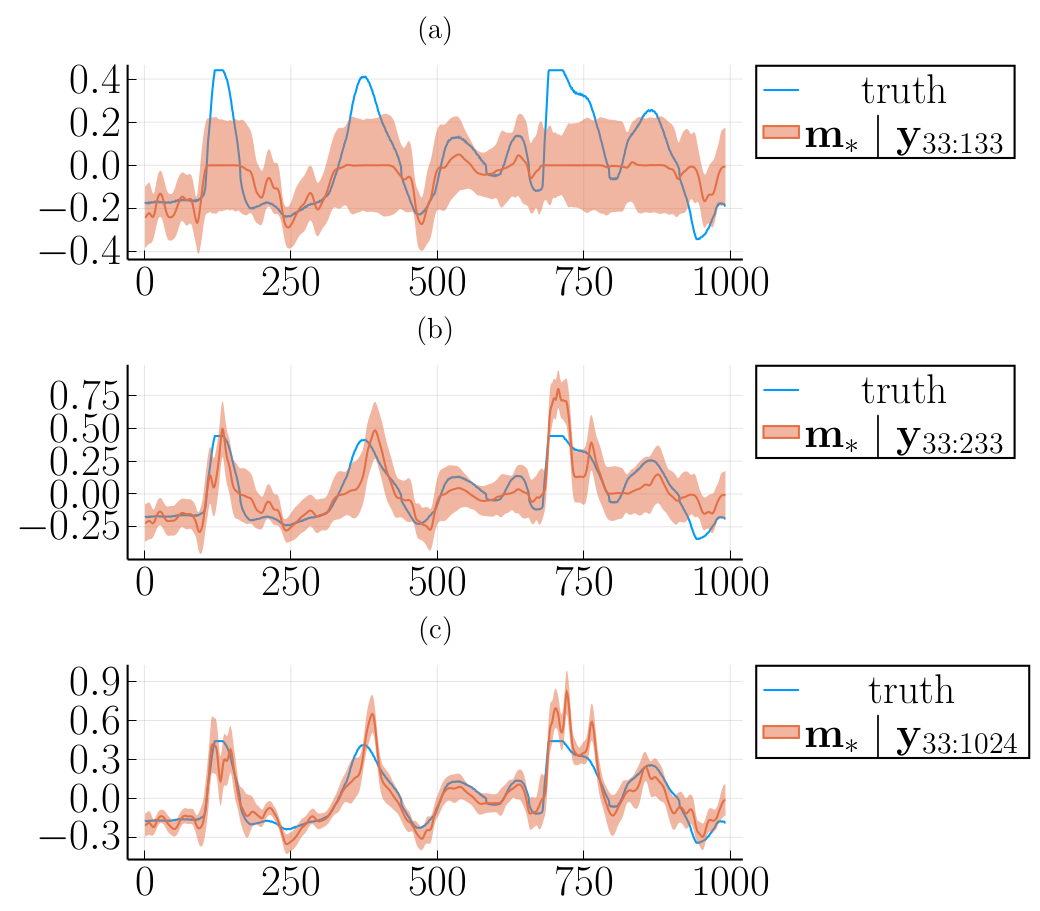}
   \caption{Predictions on test data with uncertainty bounds after seeing (a) 100, (b) 200, and (3) 992 data points for $R_\Ll=4$. The measurements start at 33 because the memory goes back 32 time steps.}
   \label{fig:predictions_3times}
\end{figure}

Finally, Fig. \ref{fig:predictions_3times} shows the predictions with the TNSRKF on testing data after seeing 100, 200, and 922 data points.
Aligned with the plot showing the RMSE and NLL, after 100 data points, the prediction is quite bad and uncertain.
After 200 data points, the predictions are better and more certain and further improve after seeing the entire data set.

\section{Conclusion}
In this paper, we presented a TT-based solution for online GP regression in terms of an SRKF.
In our experiments, we show that our method is scalable to a high number of input dimensions at a reasonable computational cost such that all experiments could be run on a conventional laptop.
In addition, we improve the state-of-the-art method for TN-based Kalman filter: In settings where the TNKF loses positive (semi-)definiteness and becomes numerically unstable, our method avoids this issue because we compute the square root covariance factors instead of the covariance matrix.
In this way, we can choose settings for our method that achieve better accuracy than the TNKF.

A future work direction is online hyperparameter optimization.
We are looking at a truly online scenario, so future data is not available.
Thus we cannot swipe over mini-batches of data multiple times like other methods, e.g.\ \cite{schurch2020recursive}, to optimize hyperparameters.

Finally, there is still ongoing research to determine how to choose TT-ranks and TTm-ranks.
In the synthetic experiments, we showed the impact of $R_\Ll$ and $R_\w$.
Generally, the TT- and TTm-ranks need to be treated as hyperparameters.

\bibliography{main}
\bibliographystyle{plain}

\end{document}